# A comparative study of zero-shot inference with large language models and supervised modeling in breast cancer pathology classification

Madhumita Sushil, PhD[1,*], Travis Zack, MD, PhD[1,2,*], Divneet Mandair, MD[1,2,*], Zhiwei Zheng, MEng[3,#], Ahmed Wali, MEng[3,#], Yan-Ning Yu, MEng[3,#], Yuwei Quan, MEng[3,#], Atul J. Butte, MD, PhD[1,2, 4,5]

1. Bakar Computational Health Sciences Institute, University of California, San Francisco, USA
2. Helen Diller Family Comprehensive Cancer Center, University of California, San Francisco, USA
3. University of California, Berkeley, USA
4. Center for Data-driven Insights and Innovation, University of California, Office of the President, Oakland, CA, USA
5. Department of Pediatrics, University of California, San Francisco, CA, USA

* co-first authorship
# equal contribution as a single team for Master of Engineering Capstone project at UC Berkeley

**Corresponding Author:**
Madhumita Sushil, PhD
Email ID: Madhumita.Sushil@ucsf.edu
Bakar Computational Health Sciences Institute, 490 Illinois Street, Cubicle 2215, 2nd Fl, North Tower, San Francisco, CA 94143
Telephone: +1 (415)-514-1971

**Word Count:** 3319 words (excluding Tables, Figures, and Supplementary Materials)

# Abstract

## Objective

Although supervised machine learning is popular for information extraction from clinical notes, creating large annotated datasets requires extensive domain expertise and is time-consuming. Meanwhile, large language models (LLMs) have demonstrated promising transfer learning capability. In this study, we explored whether recent LLMs can reduce the need for large-scale data annotations.

## Materials and Methods

We curated a manually-labeled dataset of 769 breast cancer pathology reports, labeled with 13 categories, to compare zero-shot classification capability of the GPT-4 model and the GPT-3.5 model with supervised classification performance of three model architectures: random forests classifier, long short-term memory networks with attention (LSTM-Att), and the UCSF-BERT model.

## Results

Across all 13 tasks, the GPT-4 model performed either significantly better than or as well as the best supervised model, the LSTM-Att model (average macro F1 score of 0.83 vs. 0.75). On tasks with high imbalance between labels, the differences were more prominent. Frequent sources of GPT-4 errors included inferences from multiple samples and complex task design.

## Discussion

On complex tasks where large annotated datasets cannot be easily collected, LLMs can reduce the burden of large-scale data labeling. However, if the use of LLMs is prohibitive, the use of simpler supervised models with large annotated datasets can provide comparable results.

## Conclusions

LLMs demonstrated the potential to speed up the execution of clinical NLP studies by reducing the need for curating large annotated datasets. This may result in an increase in the utilization of NLP-based variables and outcomes in observational clinical studies.

# Introduction

Over the last decade, supervised machine learning methods have been the most popular technique for information extraction from clinical notes[1]. However, supervised learning for clinical text is arduous, requiring curation of large domain-specific datasets, interdisciplinary collaborations to design and execute standardized annotation schema, and significant time from multiple domain experts for the meticulous task of data annotation. Supervised modeling can often require subsequent iterative development driven by advanced technical expertise, which can be limiting for certain practitioners. The entire process thus takes a significant amount of time between problem conception and obtaining final results. These challenges, combined with the limited availability of clinical notes corpora, have contributed to an under-utilization of Natural Language Processing (NLP) in observational studies from Electronic Health Records (EHRs)[2].

Recently, language models have demonstrated promising transfer learning ability, which is encouraging for information extraction from text without extensive task-specific model training[3–5]. Prompt-based inference is popular with generative language models, where practitioners can simply query the model in natural language to obtain the desired information, sometimes by presenting a few examples of the task they may be trying to solve. Prompt-based inference with large language models (LLMs) like the GPT-4 model have demonstrated varying levels of proficiency in medical inference tasks, such as diagnosing complex clinical cases[6–8], radiology report interpretation[9,10], clinical notes-based patient phenotyping[11,12], automated clinical trial matching[13,14], and improving patient interaction with health systems[15]. In this study, utilizing a large corpus of breast cancer pathology notes, we investigate whether large language models can alleviate the need to curate large training datasets for supervised learning to extract information from pathology text. To this end, we have a three-fold contribution:

1. We developed an annotation schema and detailed guidelines to create an expert-annotated dataset of 769 breast cancer pathology reports with document-level, treatment-relevant information. We further analyzed the curation process to identify frequent modes of disagreements in data annotation, which we additionally present here.
2. To establish a baseline of automated breast cancer pathology classification against that of expert clinicians, using the newly curated dataset, we benchmarked the performance of supervised machine learning models of varied levels of complexity, which include a random forest classifier, a long short-term memory network (LSTM) classifier, and a transformers-based medically-trained BERT classifier.
3. We finally prompted the GPT-4 and GPT-3.5-turbo models to obtain *zero-shot* classification results, i.e., results without using any domain-specific manually labeled dataset, which we compared to the supervised learning performance obtained earlier.

# Materials and Methods

**Data**

Breast cancer pathology reports between 2012 and March 2021 were retrieved from the University of California, San Francisco (UCSF) clinical data warehouse, already deidentified and date-shifted with the

Philter algorithm[16]. Breast cancer patients were identified by querying for encounters with the ICD-9 codes 174, 175, 233.0, or V10.3, or the ICD-10 codes C50, D05, or Z85.3. The cohort was restricted to pathology reports by selecting the note type 'Pathology and Cytology'. Notes shorter than 300 characters in length, and those unrelated to breast cancer, for example, those about regular cervical cancer screening through pap smears, were removed through keyword search. A flow diagram for the inclusion and exclusion criteria is presented in **Figure 1**. Among the final set of notes, 769 pathology reports were randomly selected for manual labeling with treatment-relevant breast cancer pathology.

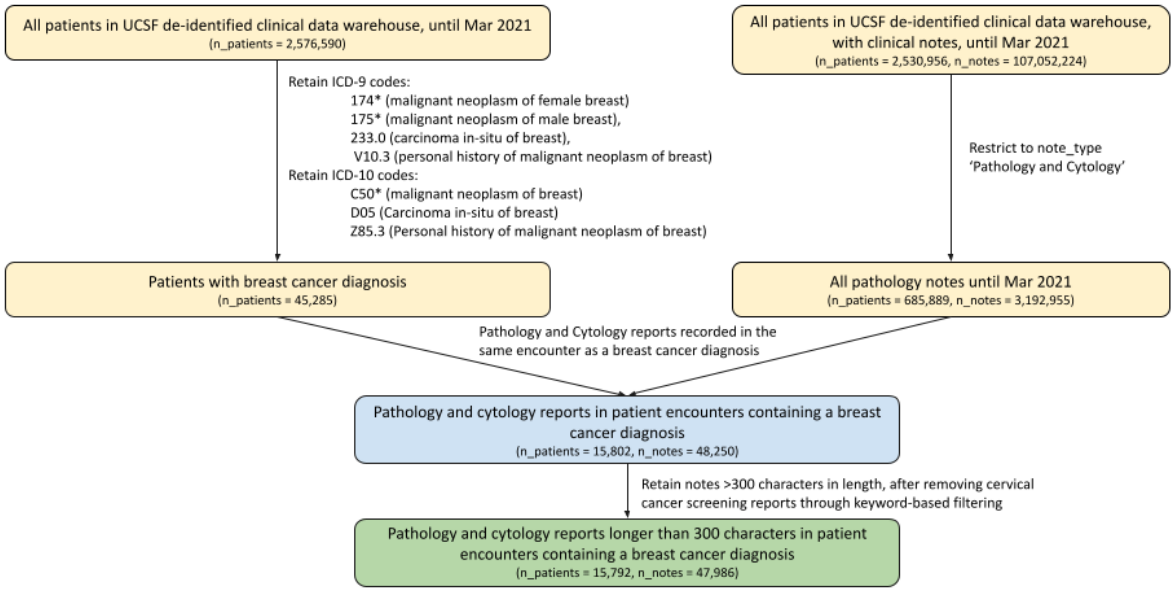

**Figure 1:** Flow diagram representing inclusion and exclusion criteria for breast cancer pathology report selection before data annotation. Number of patients and number of clinical notes is represented at each stage. The final annotated subset represents a random sample of the final representative dataset obtained in this manner.

Annotation schema and guidelines were designed in collaboration with oncology experts to label treatment-relevant breast cancer pathology details. To align with the clinical decision-making process, focus was established on identifying the most aggressive, or "worst" specimen before making document-level inferences. To analyze categories relevant for prognostic inference, categories such as final tumor margins and lymphovascular invasion were added in addition to commonly investigated categories of biomarkers, histopathology, and grade. The final cohort of 769 breast cancer pathology reports was annotated through thirteen key tasks, including ten single-label tasks and three multi-label tasks (**Figure 2**). Each report mentioned metadata such as the report date and patient ID, along with the pathologist's comments and the complete clinical diagnosis. Text spans corresponding to cancer stage, lymph node involvement, and tumor-related information were pre-highlighted within text with an external long-short term networks model that had been previously trained for named entity recognition. To establish a good inter-annotator agreement, a group of 2–3 independent annotators, either medical students or oncology fellows, annotated the documents jointly in the first phase. After achieving high inter-annotator agreement, the annotators further labeled the documents independently. The validation subset (99 reports) and test subset (100 reports) were established with documents that were at least jointly annotated, and any disagreements between discordant labels were manually adjudicated. The complete annotation guidelines are provided in the Supplementary Materials (**Section S1**).

**Zero-shot inference with LLMs**

Two large language models, the GPT-3.5 model and the GPT-4 model[17], were queried via the HIPAA-compliant Azure OpenAI Studio[1] to provide the requested category of breast cancer pathology information from a given pathology report. No data was permanently transferred to or stored by either OpenAI or Microsoft for any purposes, similar to as previously described[2]. Model inputs were provided in the format *{system role description} {note section text, prompt}*. The specific prompt, model version, and the model hyperparameters are provided in the Supplementary Materials, section S2. All thirteen classification labels were requested through a single prompt, as one call to the model for each pathology report. Prompt development was performed on the development set, and the final results were reported on a held-out test set. Model outputs were requested in the JSON format, which were post-processed into python dictionaries to automatically evaluate model outputs.

**Supervised Modeling**

Supervised machine learning classifiers were trained independently for each of the 13 breast cancer pathology classification tasks. Three models of varied complexity were included in the analysis — a random forests classifier[18], a Long Short Term Memory networks (LSTM) classifier with attention[19,20], and a fine-tuned UCSF-BERT (base) model[21,22]. The random forests model was initialized with a TF-IDF vector of n-grams within pathology notes, and the word embeddings in the LSTM model were initialized with fasttext[23] embeddings of 250 dimensions, trained on a corpus of 110 million clinical notes at UCSF. The UCSF-BERT model was pretrained from scratch on 75 million clinical notes at UCSF, and fine-tuned further on pathology classification-specific tasks. Pathology reports were pre-processed to remove punctuations and symbols, and were converted to lowercase before vectorization for the random forests and the LSTM models. For random forests single-label tasks, training data samples of the minority classes were up-sampled to reflect a uniform distribution and address data imbalance. Validation and test data were not modified and reflected the real-world distribution. To find the best parameters for the random forest model, a random grid search was performed, using 3-fold cross-validation on the training data and 15 iterations. For deep learning classifiers, the validation subset was used for hyperparameter tuning. To address the data imbalance in multi-label tasks, asymmetric loss[24] was used by the deep learning models during model training, and the categorical cross-entropy loss was used for single-label tasks. Further details on model settings and hyperparameter tuning are available in the Supplementary Materials, section S3. Source code for model implementation is further available through the github repository: https://github.com/MadhumitaSushil/BreastCaPathClassification. To obtain a reliable estimate of minority class performance, model performance was evaluated on a held-out test set with the metric macro-averaged F1-score instead of accuracy or micro-averaged F1 score.

---

[1] The AI framework is called Versa at UCSF.
[2] https://physionet.org/news/post/415

# Results

**Breast cancer pathology information extraction dataset**

769 breast cancer pathology reports were annotated with detailed breast cancer pathology information across 13 key tasks (**Figure 2**). Minimum, maximum, mean and median document length of the dataset were 36, 4430, 723.4, and 560 words respectively, and the inter-quartile range was 508 words. The dataset included a diverse population across demographics and age, with nearly 1% of cases being male breast cancer, which reflects the relative incidence of this disease (**Table 1**). Median patient age was 55 years. To encourage reproducibility and further research, upon manuscript publication, the dataset will be freely shared through the controlled-access repository PhysioNet. Average inter-annotator agreement, as quantified with Krippendorf's alpha[3][25], was 0.85 (**Table 2)**, with variability across tasks. Classification of *DCIS margins* and the multi-label category of *sites examined* showed the most interannotator discordance, while *lympho-vascular invasion* and *invasive carcinoma margin status* showed the highest concordance.

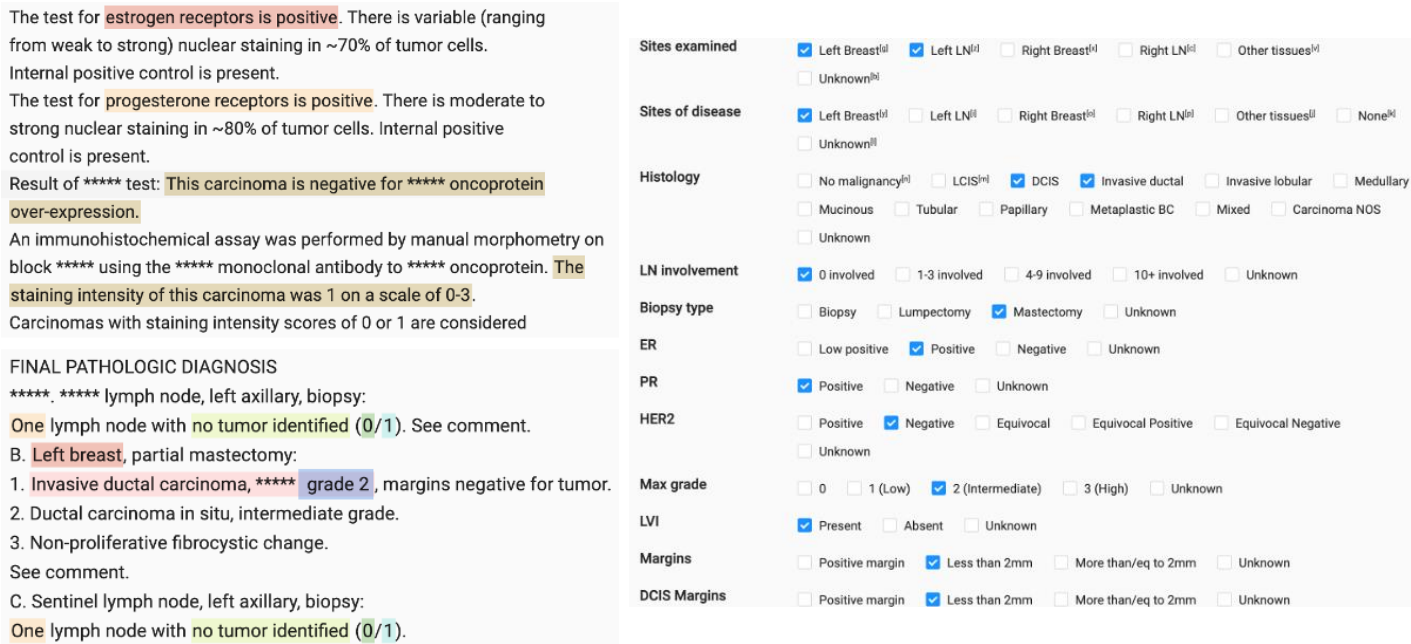

**Figure 2:** Sample of an annotated pathology report, along with the corresponding document-level annotation schema. *Irrelevant note* refers to those that are not related to a breast cancer diagnosis (further details in the annotation guidelines). The *Unknown* labels refer to the cases where a label could not be inferred based on the information provided in the pathology report.

**Table 1. Socio-demographic distribution of patients in the annotated dataset**

---

[3]Krippendorf's alpha was preferred over Cohen's kappa to quantify inter-annotator agreement on multi-label annotation tasks in addition to single-label tasks.

| Sample characteristic | Count (percentage) (n=769) | |
|---|---|---|
| Gender | | |
| Male | 7 | (0.91%) |
| Female | 762 | (99.09%) |
| Age | | |
| Median [IQR] | 55.0 | [19.0] |
| Race/ethnicity | | |
| White | 505 | (65.67%) |
| Asian | 101 | (13.13%) |
| Latinx | 42 | (5.46%) |
| Black or African-American | 36 | (4.68%) |
| Native Hawaiian or Other Pacific Islander | 7 | (0.91%) |
| Southwest Asian and North African | 2 | (0.26%) |
| Other | 23 | (2.99%) |
| Multi-Race/Ethnicity | 15 | (1.95%) |
| Unknown/Declined | 38 | (4.94%) |
| Language | | |
| English | 702 | (91.29%) |
| Russian | 18 | (2.34%) |
| Unknown/Declined | 17 | (2.21%) |
| Chinese - Cantonese | 9 | (1.17%) |
| Spanish | 9 | (1.17%) |
| Vietnamese | 4 | (0.52%) |
| Chinese - Mandarin | 2 | (0.26%) |
| Burmese | 1 | (0.13%) |
| Italian | 1 | (0.13%) |
| Cambodian | 1 | (0.13%) |
| Samoan | 1 | (0.13%) |
| Korean | 1 | (0.13%) |
| Farsi | 1 | (0.13%) |
| Sign Language | 1 | (0.13%) |
| Other | 1 | (0.13%) |

Abbreviation: IQR, Inter-quartile range

**Table 2:** Inter-annotator agreement between expert clinical annotators, as quantified by Krippendorf's alpha score for single-label and multilabel tasks.

| Task | Inter-annotator agreement (Krippendorf's alpha) |
|---|---|
| Biopsy type | 0.80 |
| Num lymph nodes involved | 0.89 |
| ER | 0.85 |
| PR | 0.90 |
| HER2 | 0.80 |
| Grade | 0.85 |
| LVI | 0.97 |
| Margins | 0.93 |
| DCIS Margins | 0.77 |
| Histology (Multilabel) | 0.82 |
| Sites examined (Multilabel) | 0.79 |
| Sites of disease (Multilabel) | 0.85 |
| **AVERAGE** | **0.85** |

Sources of disagreements between annotators in the development and the test sets were analyzed by an independent adjudicator. Common sources of disagreements included differences in inferring the most aggressive ("worst") sample when multiple samples were analyzed, incorrectly including information from patient history for providing labels for the current report, linguistic or clinical ambiguity in the pathology report, discordant interpretation of procedures involving excisions when differentiating between a histopathology report and a cytology report, inconsistencies in categorizing metastatic disease sites as "other tissues" and histology as "others", and inconsistent execution of the annotation guidelines for annotating molecular pathology reports and grade information.

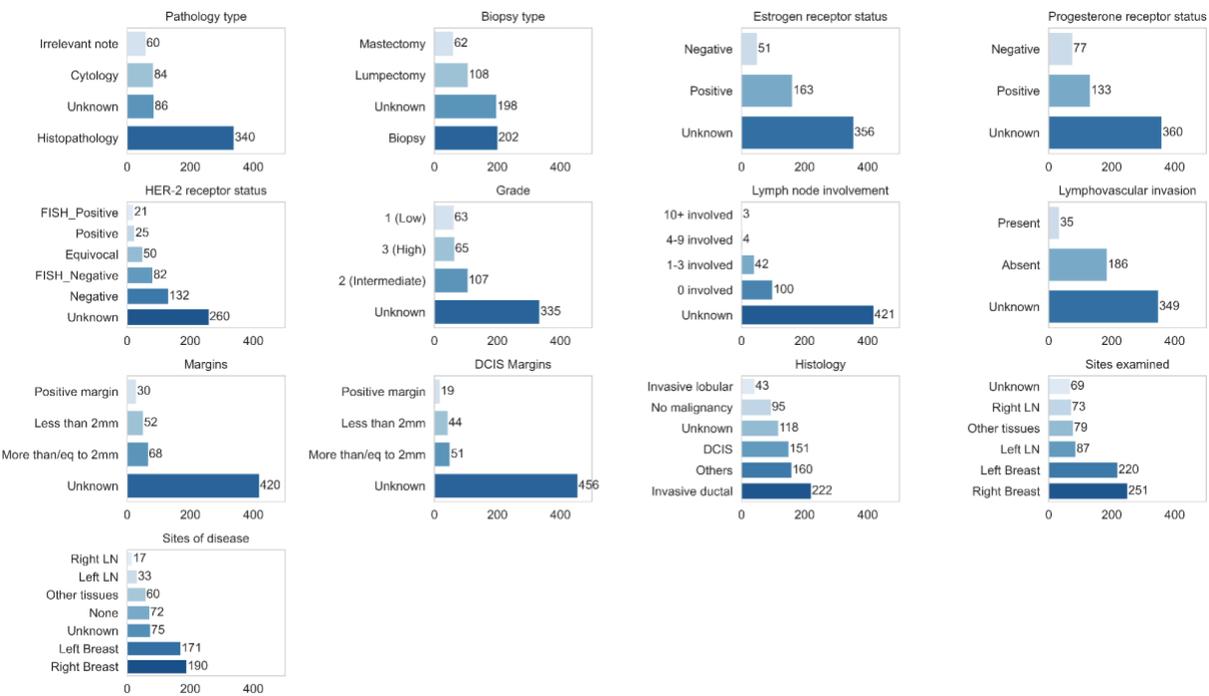

**Figure 3:** Class distribution for all tasks in the training data for supervised classification.

The class distribution across the annotated data was highly skewed, resulting in a highly imbalanced dataset. Certain low frequency categories, such as groups of histology codes or the *low positive* and *positive* categories of Estrogen Receptor status were combined before further automated classification, resulting in the final distribution presented in **Figure 3**. The *Unknown* class, which corresponded to the case where the requested information could not be inferred from the given note, was the majority class across 8 of 13 tasks. Among the remaining classes, high imbalances were observed in tasks of inferring the category of the number of lymph nodes involved, lymphovascular invasion, tumor margins, and HER-2 receptor status.

## The GPT-4 model is as good as or better than supervised models in breast cancer pathology classification

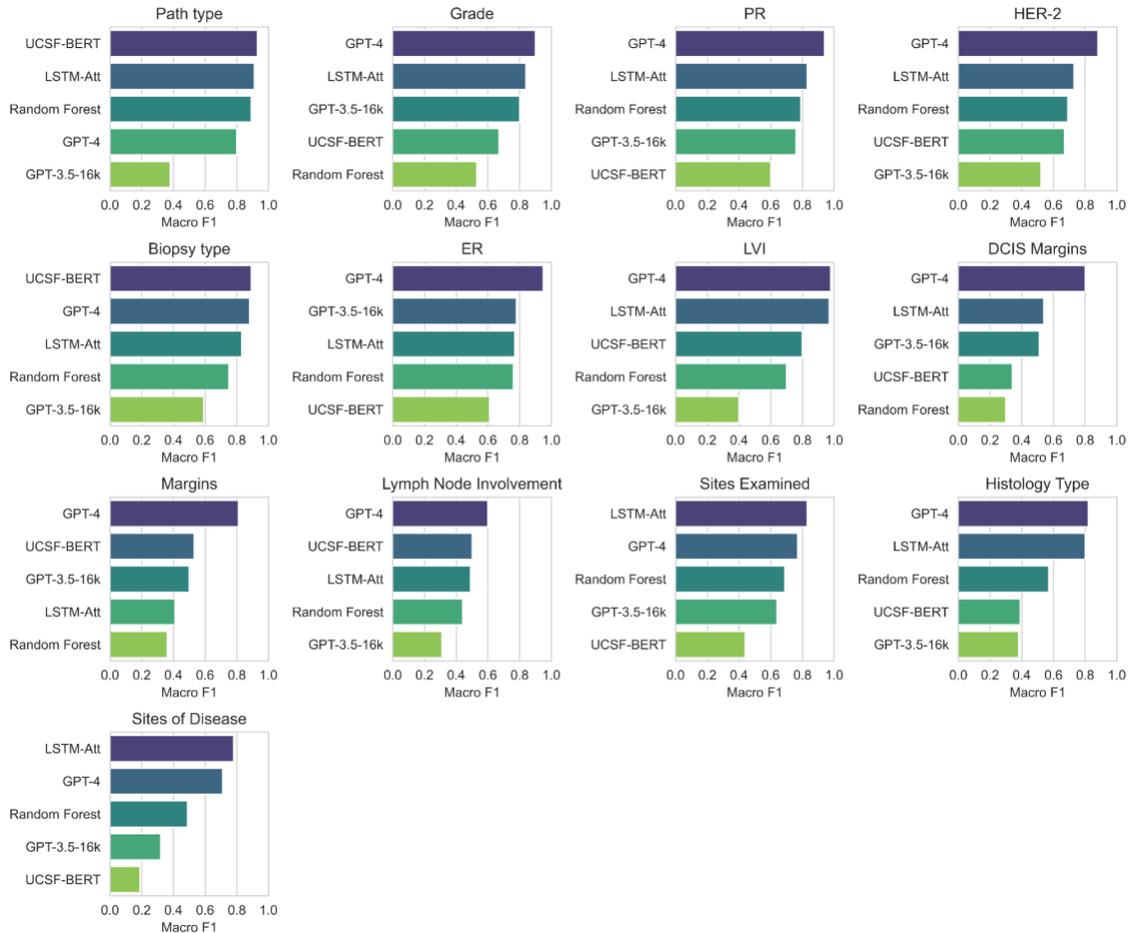

**Figure 4:** Classification performance, as measured by Macro F1, for different models for each classification task. All models other than GPT-3.5 and GPT-4 are trained in a supervised setup on task-specific training data. GPT-3.5 and GPT-4 models are evaluated zero-shot, i.e., in an unsupervised manner.

Despite no task-specific training, the GPT-4 model either outperformed or performed as well as our tasks-specific supervised models trained on task-specific breast cancer pathology data (**Figure 4**). For both the GPT-4 model and the GPT-3.5 model, all model responses could be automatically parsed as JSON without any errors. The average macro F1 score of the GPT-4 model across all tasks was 0.83, of the LSTM model with attention was 0.75, of the random forests model was 0.61, of the UCSF-BERT model was 0.58, and that of the GPT-3.5-turbo model (zero-shot) was 0.53. The GPT-4 model was significantly better than the LSTM model (the best supervised model) for the tasks of margins and estrogen receptor

(ER) status classification (Approximate randomized testing for significance[26], p < 0.01). These tasks encompass either a large training data imbalance resulting in a sparsity of class-specific training instances (margins) or could be frequently solved with an n-gram-matching approach (ER status). For all other tasks, no significant differences were obtained between the zero-shot GPT-4 model and the supervised LSTM model. The GPT-3.5-turbo model performed significantly worse than the GPT-4 model for all tasks. Similarly, the UCSF-BERT model, which is a transformers model pre-trained on the UCSF notes corpus[21,22], did not outperform simpler models like random forests or LSTM with attention for several tasks, potentially due to relatively small sample sizes and highly imbalanced training data. The random forests classifier performed well on keyword-oriented tasks, like pathology type classification and biomarker status classification, but under-performed on tasks requiring more advanced reasoning, like grade and margins inference.

**GPT-4 model error analysis**

As demonstrated in **Figure 5**, the confusion matrix of the GPT-4 model revealed that it had difficulties in differentiating the unknown class from the class that indicated no lymph node involvement and no lympho-vascular invasion. Furthermore, margins inference was complex for the model, where *more than 2mm margins* (negative margins) are confused with *less than 2mm* margins. Confusion between classes were more prevalent in multi-label tasks than single-label tasks. Further errors from the GPT-4 model were prevalent when the task design was ambiguous in model prompts, such as the grouping of sparse histology into an "others" category, the assignment of metastatic sites for breast cancer as "other tissues than breast or lymph nodes", or the inference of pathology reports unrelated to breast cancer. The latter set of errors correspond to common sources of disagreements identified during the data annotation process.

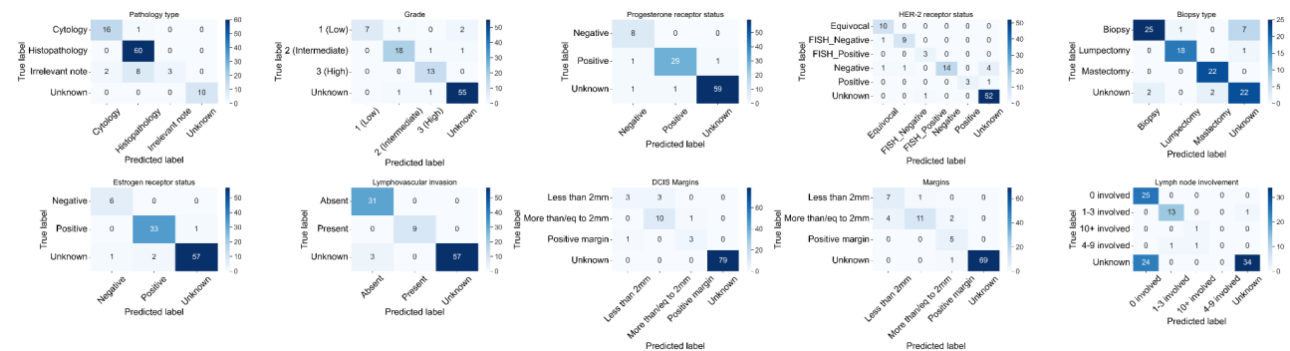

(a) Single-label classification tasks

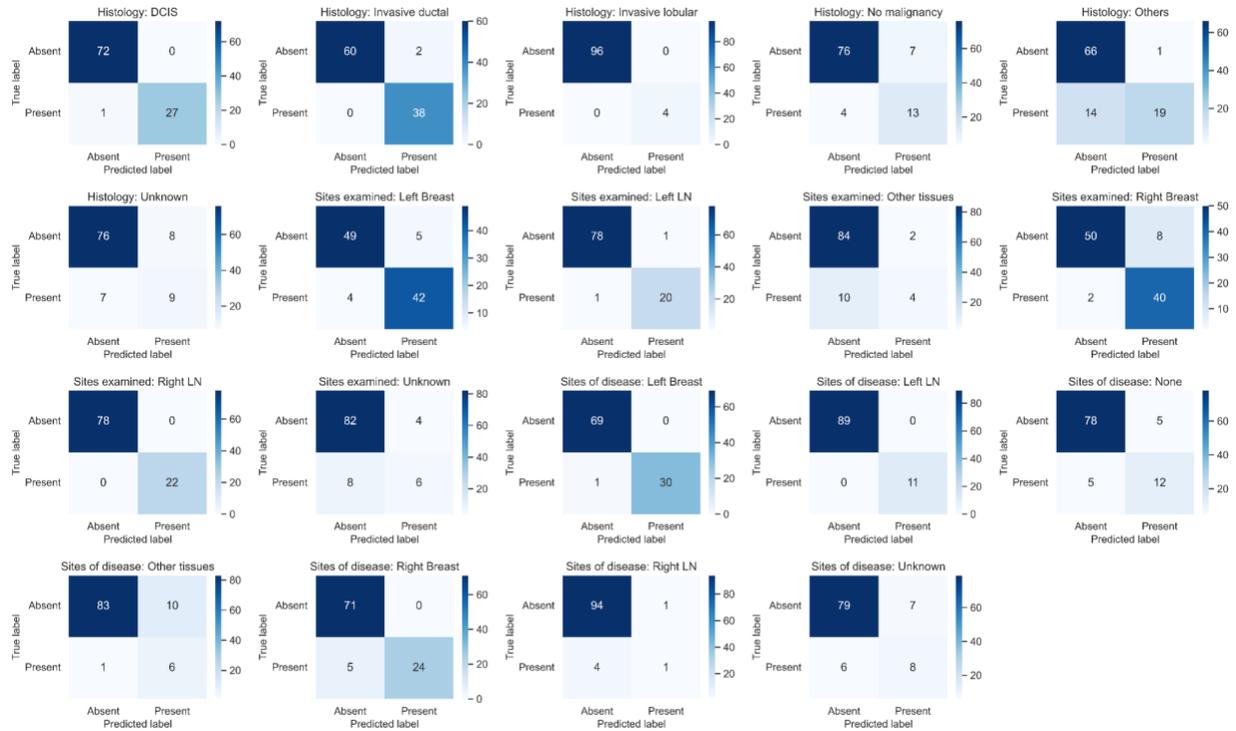

(b) Multi-label classification tasks

**Figure 5:** Confusion matrices for GPT-4 classification in (a) single-labeled tasks, (b) multi-labeled tasks.

Manual analysis of the GPT-4 model errors revealed several consistent sources of errors. Common sources of errors in biomarker reporting involved the reporting of results from clinical history or tests conducted at other sites that were not confirmed in the current report. Furthermore, the GPT-4 model incorrectly reported nuclear grade as the overall tumor grade when the overall grade was not discussed in the note. Moreover, common errors in reporting tumor margins were concerned with mathematical inferences over multiple margins (for example, anterior, posterior, medial, etc.), where the category representing the closest margin values needed to be provided. Manual analysis additionally uncovered several error sources for multi-label tasks. The model performed inconsistently when inferring sites of benign findings; while the model frequently missed reporting the site of benign findings as a site examined for tumors, it also sometimes included sites of benign findings as a site of cancer. Furthermore, sentinel and axillary lymph nodes were frequently reported as tissues other than breast or lymph nodes, although they were annotated as lymph node sites. Some errors related to complex cases were also found, for example, 1% staining results for progesterone receptors were provided as negative by the model, whereas they were annotated as positive. Finally, errors related to task setup were reflected in histology-related errors, where the model could not reliably abstain from providing histology from reports unrelated to breast cancer and from molecular pathology reports for ERBB2 despite being instructed as such, and errors due to the grouping of histologies like LCIS into an "others" category.

# Discussion

Task-specific supervised learning models trained on manually annotated data have been the standard approach in clinical NLP for over a decade[1]. Using a manually annotated dataset of 769 breast cancer pathology reports focused on the most clinically relevant report features, our study compared the performance of supervised learning models, including random forests classifier, LSTM models, and the UCSF-BERT model, with a zero-shot classification performance of two LLMs, the GPT-4 model, and the GPT-3.5-turbo model. We found that even in zero-shot setups, the GPT-4 model performs as well as or significantly better than simpler, task-specific supervised counterparts, although the GPT-3.5 model performs significantly worse than the GPT-4 model on all classification tasks. Previous studies have demonstrated similar results, showing that in zero-shot setups, LLMs consistently perform the same as or outperform fine-tuned models on biomedical NLP datasets with small training data sizes (fewer than 1000 training examples)[27,28]. Similar small datasets are common in medical informatics studies since domain expertise is frequently required for reliably annotating clinical notes, making the process time-consuming and difficult to scale[29]. This study confirms previous findings on a new real-world clinical dataset, reinforcing that LLMs are promising for use in classification tasks in low-resource clinical settings.

Tasks where the training data contained high class imbalance or keyword-based tasks (i.e., could be largely solved with simple lexical matching) were particularly conducive for using GPT-4 model over task-specific supervised models. Given that the GPT-4 model is already trained on internet-scale corpora, the model may already encode a fundamental understanding of breast cancer pathology-related terminology, which may explain its surprising zero-shot capability on these tasks, including that on complex and imbalanced tasks like margins inference. However, the reasons behind the striking performance difference between the GPT-3.5 and GPT-4 models remain unclear due to the closed nature of these models, although similar trends have been observed in previous medical NLP studies[30–32].

An analysis of the GPT-4 model errors indicated several errors due to insufficient understanding of idiosyncratic task-design choices, for example differentiating between "Unknown" and "no lymph node involvement" categories, and grouping of less frequent histologies into an "others" category. It is possible that these errors can be mitigated with strategies, such as few-shot learning to demonstrate a better understanding of annotation-specific choices, or chain-of-thought-prompting to elucidate reasoning and avoid answering from incomplete or old information within text report. However, it has been demonstrated earlier that the GPT-4 model cannot process long input contexts efficiently, particularly when the results are included in the central part of the context[33]. Hence, how to best integrate in-context learning with long pathological notes such that the model can still make effective use of the supported context length remains to be investigated in future research.

Although this study compared two proprietary LLMs with supervised classifiers on a real-world breast cancer dataset, several design choices may have impacted the findings. The dataset was curated from a single health system, and further validation of the findings on pathology reports from other health systems may improve the reliability of the results. Although potential de-identification errors may have impacted the capability of LLMs, the data reflects real-world setups for retrospective observational studies in a privacy-preserving manner. Furthermore, although some model-specific hyperparameters were explored for baseline models, all possible choices have not been explored, and it may be possible to improve baseline model performance further with continued development. Moreover, LLMs used in the study were evaluated using a single prompt and model setting, and the results could be sensitive to these design choices. However, the findings of this study will inform future studies on the development of more advanced prompting and few-shot strategies for LLMs to obtain even better performance, the development of effective annotated datasets for simpler supervised classification setups, the evaluation of newer LLMs for clinical information extraction, and the analysis of output sensitivity to input prompts and model settings. While the findings from this study demonstrate the promising capability of LLMs for clinical research, if access to models like the GPT-4 model is prohibitive due to either privacy or computational constraints, comparable performance on EHR-based NLP tasks such as pathology classification can be obtained with simpler deep learning classifiers, particularly if annotated sample sizes are sufficiently large and class imbalance can be controlled through targeted annotations of minority classes for model training. Finally, the studied classifiers may exhibit biases against specific demographics, and caution must be exercised when deploying them in clinical workflows. These biases need to be investigated further in the future to establish concrete guidelines for their use.

Despite widespread studies in oncology information extraction from textual clinical records[34,35], annotated datasets of breast cancer pathology reports are not publicly available. To make the findings of this study replicable and promote further research on breast cancer pathology extraction, the dataset curated in this study along with corresponding source code for the supervised machine learning pipelines and zero-shot LLM inference will be shared publicly through a controlled-access repository PhysioNet, accessible via a data use agreement. Providing this large, annotated dataset for future research increases the impact of this work by allowing it to serve as a baseline of comparison as LLMs continue their rapid progress.

# Conclusions

The study compared breast cancer pathology classification abilities of five models of varying sizes and architecture, finding that the GPT-4 model, even in zero-shot setups, performed similarly to or better than the LSTM model with attention trained on nearly 550 pathology report examples. The GPT-4 model outperformed simpler baselines for classification tasks with high class imbalance or that required simpler keyword-matching for inference. However, when large training datasets were available, no significant difference was observed between the performance of simpler models like the LSTM model with attention compared to the GPT-4 model. The results of this study demonstrated that while LLMs may relieve the need for resource-intensive data annotations for creating large training datasets in medicine, if there are privacy, computational, or cost-related concerns regarding the use of LLMs with patient data, it may be possible to obtain reliable performance with simpler models by developing large annotated datasets, with particular focus on minority class labeling potentially in an active-learning setup.

# Acknowledgements

This research would not have been possible without support from several people. The authors thank the UCSF AI Tiger Team, Academic Research Services, Research Information Technology, and the Chancellor's Task Force for Generative AI for their software development, analytical and technical support related to the use of Versa API gateway (the UCSF secure implementation of large language models and generative AI via API gateway), Versa chat (the chat user interface), and related data asset and services. We are grateful to Dima (Dmytro) Lituiev for sharing LSTM-based token-level pathology information extraction pipeline for pre-annotating breast cancer pathology reports. We thank Boris Oskotsky, the Information Commons team, and the Wynton high-performance computing platform team at UCSF for supporting high performance computing platforms that enable the use of language models with de-identified patient data. We further thank Debajoyti Datta and Michelle Turski for feedback on breast cancer pathology annotation schema, all members of the Butte lab for helpful discussions in the internal presentations, and Gundolf Schenk and Lakshmi Radhakrishnan for discussions related to clinical notes de-identification. Partial funding for this work is through the FDA grant U01FD005978 to the UCSF–Stanford Center of Excellence in Regulatory Sciences and Innovation (CERSI), through the NIH UL1 TR001872 grant to UCSF CTSI, through the National Cancer Institute of the National Institutes of Health under Award Number P30CA082103, and from a philanthropic gift from Priscilla Chan and Mark Zuckerberg. The content is solely the responsibility of the authors and does not necessarily represent the official views of the National Institutes of Health.

# Supplementary Materials

## Section S1: Annotation guidelines

**Pathology Type**

Path type   ☑ Histopathology[s]   ☐ Cytology[d]   ☐ Irrelevant note[f]   ☐ Unknown[g]

This is a single-option selection regarding the TYPE of sample a pathology report is referring to. In general, there are two main types of samples in pathology.

1. Cytology (Cyto="cell"). In this type, cells are collected from either fluid or from a procedure called "Aspiration". In either of these two techniques, the structure BETWEEN cells is lost, so the only information gathered are what the individual cells look like under the microscope
2. Histopathology: (histo="tissue"): as the name suggests, a sample is collected as an entire "Chunk" of tissue. This allows for characterization of both cellular features but ALSO the structure and organization of those cells within the larger tissue/organ/tumor.
3. Irrelevant: In some cases, we have collected pathology notes on breast cancer patients that are not RELATED to cancer diagnosis or treatment directly. For example notes about reconstruction surgeries, or perhaps any notes related to cervical cancer screening pap smears. In these cases, please mark notes as Irrelevant
4. Unknown: In some cases, it can be very difficult to determine if the primary sample is cytology or pathology. The most common example of this is for Molecular staining for a specific tumor marker (in our case, specifically for HER2/ERBB2), where the pathologist is only looking for a single presence or absence. In these cases, please always mark Unknown.

    The pathology type can generally be determined at the beginning of a report.

    **Examples of Histopathology key words:**
    - *Surgical Pathology report*
    - *Lumpectomy*
    - *Core biopsy*
    - *Excisional biopsy*
    - *Mastectomy*
    - Additionally, mention of specific forms of carcinoma require histopathology for diagnosis, so if "adenocarcinoma", ductal carcinoma, etc are described, this is a histopathology note

    **Examples of cytology key words:**
    - Fine Needle Aspiration
    - Cytology
    - XXXX of XXX *fluid*

    **Examples of Irrelevant notes**
    - Capsulectomy or breast reconstructive surgery
    - Cytology report for pap smear in patient with history of breast cancer

**Disambiguation**

In some notes, multiple tumor samples may be present (such as surgical reports). In these cases where both histopathology and cytology reports are present, please label as histopathology.

**Sites examined and Sites of disease**

| Sites examined | ☑ Left Breast[z] | ☐ Left LN[x] | ☐ Right Breast[c] | ☐ Right LN[v] | ☐ Other tissues[b] | ☐ Unknown[y] |
|---|---|---|---|---|---|---|
| Sites of disease | ☐ Left Breast[i] | ☐ Left LN[o] | ☐ Right Breast[p] | ☐ Right LN[j] | ☐ Other tissues[k] | ☑ None[l] |
| | ☐ Unknown[n] | | | | | |

This is a **multi-class selection for which tissues are present in the report and which tissues contained malignancy**.

Often, much of this information can be gather at the beginning of the report under the section: Final Pathological Diagnosis:

FINAL PATHOLOGIC DIAGNOSIS
A. Left breast, wire-localized lumpectomy:
1. Microcalcifications in benign breast tissue; see comment.
2. Prior surgical site changes.
B. Left breast, anterior margin, excision: Benign breast tissue with duct ectasia; see comment.

For sites examined, please check all sites in which there is a specimen that was examined for disease. In the above example, *left breast* would be checked, even though both specimens were benign. Please make the best guess when the laterality is not 100% clear.

For sites of disease, check all boxes where any type of disease has occurred in that site (this includes, DCIS, invasive cancer, etc.).

Sites of disease is *None* when we know that tumors were not found among the sites that were examined. Unknown is to be selected only when we can't make the inference from the data, for example, when the results are not present in the report.

Any lymph nodes, for example supraclavicular lymph node, should be considered as lymph nodes and not "other tissues". Finally, we often find that skin is one of the specimens examined. If the specimen is referring to skin overlying breast tissue, then this should NOT be labeled as 'other tissue'. Instead, skin overlying breast tissue should just be grouped with the same labeling for the underlying breast tissue. The reason for this is that for breast cancer, if the skin overlying a breast also has evidence of disease, then this reflects the degree of LOCAL INVASION of that cancer (ie it would have a T stage of 4). So we really don't want to count this as a separate 'other tissue' - the skin involvement would be reflected in the staging information we capture about the disease (it's not a separate site of disease). You would only check 'other tissue' for skin involvement if the site of skin involvement is some other area of the body, for instance the arm, leg, back, etc.

There is also a text span "Tumor location" closely associated with this label. If an invasive disease is present, you should highlight the text that corresponds to where this is located ONLY (do not highlight locations of DCIS or benign samples if invasive disease is present). If only DCIS is present, then you should highlight the text that corresponds to where the DCIS is located. Only if all specimens are benign should you highlight the locations of benign disease with 'tumor location.' In the text span, you should include not only laterality (ie Left Breast) but also positional information often included alongside this, e.g. '10 o'clock position'. You should only need to label tumor location text spans in one of 2 sections in the pathology note: in the 'FINAL PATHOLOGIC DIAGNOSIS' and the 'Comments' sections.

**Histology**

| Histology | ☑ No malignancy[m] ☐ LCIS ☐ DCIS ☐ Invasive ductal ☐ Invasive lobular ☐ Medullary ☐ Mucinous ☐ Tubular ☐ Papillary ☐ Metaplastic BC ☐ Cribiform ☐ Mixed ☐ Carcinoma NOS ☐ Unknown |
|---|---|

Histology types - This is a **multiclass selection for all histology types that correspond to the specimens you've labeled in the text span**. If there are multiple tumor specimens, at the document level we are annotating the 'worst' of these for both invasive tumors AND DCIS. More below on how you decide what's the 'worst' invasive tumor. But the point to remember here is that you should select the histology that corresponds to that worst case and if DCIS is present in any of the specimens, you should also check that (DCIS still has significance if it occurs along with invasive tumors, but more on that below). 'No malignancy' should be selected only when there is no disease of any kind found in any of the samples (note that in these cases, you should also have 'None' checked for sites of disease). Note that the category 'Cribiform' has been removed moving forward, and can be ignored in the guidelines. The category 'Others'[4] is reserved for cases where a histology is present, but it is not present in the options provided by us in the 'Histology' class list.

For text spans, the histo type corresponds to the text span 'Tumor type'. If an invasive tumor is present, the text span highlighted should be for the corresponding histo type. Do not highlight benign samples as 'tumor type' if invasive disease (or DCIS) is present. If DCIS is present, it also should be highlighted as 'tumor_type'. Only if there are no invasive samples nor DCIS in a path report should you highlight benign samples that mention no tumor as 'tumor type'.

The histology almost always should be annotated from the "Final Pathological DIagnosis section" - many times it is duplicated within more technical portions of the text. It is not necessary to mark all the text spans where the histology is mentioned if it is the same as what is contained in the 'Final Pathological Diagnosis' Section. Even if you initially think it's clear that a specimen is lobular invasive from the 'Final Pathology Diagnosis section", you should quickly look over the path description to make sure no other histology types were mentioned (you don't have to be exhaustive about this, but a quick check here might show the pathologist also saw some ductal components, for example). The key histology types to really focus on are LCIS, DCIS, invasive ductal and invasive lobular, as these are by far the most frequent types of in situ and invasive carcinomas for breast
cancer.

"Mixed" refers to a situation where ~50% of the cells are one type but there is a second population that makes up >10% of the remaining cells: https://www.pathologyoutlines.com/topic/breastmixedNST.html.

---

[4]Please note that in new iterations, Carcinoma NOS has been changed to an 'Others' category to represent all the tumors that don't fit the remaining category.

This description is often contained in a separate descriptive paragraph. If the sample is described as mixed in that paragraph, typically it would be described as mixed in the Final Path diagnosis. In general, we decided to select both if both are present in cases as neither is necessary "more" aggressive than the other. We can leave out invasive ductal and invasive lobular individually if these refer to Mixed carcinoma in the note.

**Lymph Node Involvement**

This includes the total number of lymph nodes that are involved (including left and right sides, e.g. for bilateral mastectomy). If there are multiple samples from the same date, we should sum those lymph node numbers. If there are multiple samples from different dates, then the maximum number of lymph node should be selected independently (this scenario becomes a bit tricky as you need to monitor sample dates). Any lymph node, for example supraclavicular lymph node, should be considered as lymph nodes and not "other tissues".

Unknown should be marked if the lymph nodes were not examined at all in the given pathology report. If the lymph nodes were examined, but disease was not found on lymph nodes, the '0 involved' category should be marked. It might feel redundant (which means you're annotating correctly :)) to check '0' lymph nodes involved when you've already checked 'None' for sites of disease, but that's just the way it goes. Please make the best guess when the number of lymph nodes involved is not 100% clear.

Some technical pointers: if the number of lymph nodes with disease is not clear (whether this is due to redaction or the note simply doesn't mention), then either make the best guess or check 'Unknown'.

The corresponding text span-level annotations of 'tumor_location' will be highlighted only if a tumor was found in a lymph node.

**Biopsy Type**

| Biopsy type | ☑ Biopsy | ☐ Lumpectomy | ☐ Mastectomy | ☐ Unknown |
|---|---|---|---|---|

This is a single item selection. Sometimes, a path report will refer to multiple specimens. These may be specimens from different points over time or different sites where tissue was examined. Key things to remember when annotating the biopsy type (a label that's only at the document level), is that this should correspond to the specimen you've chosen as the 'worst' disease. However, if all the samples are disease-free, just mark the biopsy type to be the highest level of sample analyzed.

More technical things to remember are that 'biopsy' here refers to Fine Needle Aspiration (FNA) as well as core needle biopsies.

Lumpectomies should be checked for any surgery with partial removal of breast tissue - in most cases, the note will call the procedure a lumpectomy and this will be clear. Some notes, however, refer to lumpectomies as 'excisional biopsies or re-excision biopsy' - this will take a bit more detective work to confirm but this generally is still referring to a lumpectomy, since portions of breast tissue were removed but the specimen does not have true margins (which you'd only get with a mastectomy).

For non breast tissue, some general rules are that Fine Needle Aspiration (FNA) or procedures involving needle-like instruments should be grouped under biopsy: these include CSF or other percutaneous biopsies. Other times a procedure may not fit nicely into our groupings (such as excision of a metastatic brain tumor) and 'Unknown' should be checked.

If mastectomy has been mentioned only in the history or only in reference to a cosmetic surgery, those should be ignored.

Refer to the end of the document to read about how to choose the appropriate 'worst' invasive disease. The same logic applies to grade, margins, ER/PR status below as well.

---

**Grade**

> **Max grade**    ☐ 0   ☐ 1 (Low)   ☑ 2 (Intermediate)   ☐ 3 (High)   ☐ Unknown
>
> - Invasive tumor grade (modified *****-*****): Tumor is too small for accurate grading, at least ***** grade 2.
> Nuclear grade: High grade, 3 points.
> Mitotic count: Tumor too small for full mitotic count, at least 1 point.
> Tubule/papilla formation: No tubule formation, 3 points.
> Total points and overall grade = at least 7 points = at least grade 2.

Grade should also be annotated in two ways simultaneously: 1) as document-level class, 2) corresponding text spans should also be highlighted.

Like the document-level annotations, the grade checked is the overall grade for the specimen that you identified as the 'worst' - note, if there is any invasive disease in the report, this grade should correspond to invasive disease. If only DCIS is present in the report, then this grade would correspond to DCIS. If grade is only present in history, it should be ignored. If it is present as results from another center but not confirmed at UCSF, it should be marked.

For text spans of grade in the pathology report, we only annotate mentions of OVERALL grade for any particular specimen. There are only two sections where you should be highlighting text areas corresponding to grade (otherwise, you may be highlighting a number of random grade mentions). The 'Final Pathologic Diagnosis' section often but not always mentions the overall grade for a specimen. Secondly, as seen above, there's often a section in path notes 'Comments' that will mention the overall grade twice (as depicted above). You should highlight in the text span all of these and ignore any others. Mentions of the word 'grade' itself, then, are not included in these text spans and should be deleted if marked already. Well/moderate/poorly differentiated tumors represent grades 1/2/3 respectively. Nuclear grade is equivalent to overall grade only for DCIS and should only be highlighted in text as the grade if there is only DCIS in the path report (otherwise, the grade highlighted in the text should correspond to the invasive disease). Another point is that grade should never be annotated from cytology or Molecular HER2 study reports - often these reports might mention the nuclear grade but without tissue architecture, a true pathologic grade can't be determined. You should have 'Unknown' marked for grade in these cases.

Overall grade is often mentioned as a part of a larger text string that begins with 'Total points/overall grade' - all of this, up until the mention of the overall grade of the sample, is included in the text annotation span.

Sometimes the grade that is mentioned as a fraction is replaced with a date, for example 3/5 being represented as September 3rd instead. This can happen due to the incorrect redaction of PHI data, where the redaction algorithm assumed that a fraction is date, shifted it, and replaced it. In this case, make the best guess of the grade from the surrounding text for a document-level category and mark those spans, instead of using this incorrect information for inference. If an inference cannot be made, leave it as Unknown.

---

**ER**

| ER | ☐ Low positive | ☑ Positive | ☐ Negative | ☐ Unknown |

The test for estrogen receptors is positive. There is strong nuclear staining in >90% of tumor cells. External positive control is present.

Estrogen receptor values should be selected at both text span level as well as document-level. When annotating at the text span-level, the percentage positive should be included in the text spans when mentioned, for example "ER was 90% positive".

Any mentions of ER in the history (for a previously treated disease) should be ignored, unless it has been confirmed in the current state of the disease either at UCSF or at an external center. This is so because ER values can change over the course of the disease. However, if ER status is mentioned as finding from another center (for the current disease), it should be included in the annotations even if it has not been confirmed at UCSF.

If multiple values are present, make sure that the ER result corresponds to the same specimen you've picked as the 'worst' disease. For example, if there are two different statuses: one before a neoadjuvant therapy, and one after, then we annotate the one after.

We have two categories for positive ER disease: Low positive and positive. Low positive is for cases under 10% staining, also called "weakly positive". This differentiation has been made to account for differences in treatment guidelines for the same.

---

**PR**

| PR | ☑ Positive | ☐ Negative | ☐ Unknown |
|---|---|---|---|

> The test for <mark>progesterone receptors is positive. There is strong nuclear staining in >90% of tumor cells</mark>. External positive control is

Progesterone receptor status is annotated in the same manner as estrogen receptors. The only difference is that we do not differentiate between low positive and positive disease for the same. So even if PR status is "barely present", it should be marked as positive. Please include %age positive within the text spans for this.

If ER and PR values are mentioned within the common text, for example 'ER and PR were found to be positive", overlapping text spans should be added for both text-level labels.

---

**HER2**

| HER2 | ☐ Positive | ☐ Negative | ☐ Equivocal | ☑ Equivocal Positive | ☐ Equivocal Negative | ☐ Unknown |
|---|---|---|---|---|---|---|

> <mark>Positive: Tumor cells showing amplification of \*\*\*\*\* (ERBB2).</mark>

HER2 status is similar to the prior ER and PR, although a few specific issues arise at both the document and text level. For the document level, an original specimen will have IHC staining to test for HER2 - based on these results (if HER2 is present in the report), you should check Positive, Negative (0 or 1+) or Equivocal (2+). The note will mention the level of the stain and often explicitly mention if it's negative/positive or indeterminate. Note that if HER2 results have been reported as FISH results, they should always be marked as "Equivocal Positive", "Equivocal Negative", or "Equivocal", and not directly as "Positive or Negative". This is so because specimens that are equivocal will have follow up FISH studies for HER2.

Sometimes, these follow-up will look like a completely different style of path report (Molecular pathology report). Buried in the text in these reports, you will see a line like above, that shows the status of the FISH HER2 test (called ERBB2 here). This will either be positive or negative (there will be no indeterminate result here) - check either Equivocal Positive or Equivocal Negative for these reports. We do this since the FISH analysis was performed generally as a follow up test for equivocal cases (this is not always true but a generalizing assumption that helps us break out what type of test was performed). **For text spans, make sure you include not only if the result was positive but all the description of the test result (degree of IHC staining, etc.)**

Moreover, HER2 term itself may have been incorrectly redacted at times (example in the screenshot above). In this case, if HER2 can be inferred from the context, it should still be annotated.

Again, any mentions of HER-2 in history should be ignored.

---

**Margins**

Margin status would be "positive" if an invasive tumor touches normal tissue, "close" if tumor was <2 mm from the normal tissue, and "negative" if tumor was ≥2 mm from the normal tissue. Here, the least distance is considered to be the worst case.

There are often multiple specimens in the same path report, hence there may be multiple margins. At the note level, the margin selected is the worst margin status (i.e. positive margin if present) for the dominant/worst disease site in the note. If there have been multiple resections for a given specimen (i.e. the same invasive lobular cancer in the left breast), then the margin at the end of the final resections should be selected. This is true even if a path report has multiple specimens from the same disease/site over time.

To help with text span annotations, margin information should be obtained from two places: the "Final Pathologic Diagnosis' Section and the 'Comments' section under the specimen of interest. In the "Comments' section, there is often a detailed list of the margins for all directions of the sample (anterior/inferior, etc.). All of this should be highlighted in a text span with the label 'tumor margin'.

The document level class label for margin will reflect the closest margin status across all of these borders. i.e., if multiple specimens are present, the following priority order is established: the first choice would be positive margin if any tumor touches the normal tissue, the second choice would be < 2mm if the distance between the tumor and the normal tissue is <2mm, the third choice would be >= 2mm or negative margins. Please pay attention that you do not make mistakes in converting between mm and cm; our margin options are mentioned as mm. 1cm = 10mm, or vice versa, 1 mm = 0.1cm. Unknown should be selected only if the margin is not mentioned in the report. When the report only mentions "margin clear", "free of tumor", or "margin negative", it should be categorized into "More than/eq to 2mm".

---

**DCIS Margins**

Annotated in the same manner as Margins, but for DCIS only. Note that the DCIS margins task has a separate label for text annotations. Look for the same sections as with invasive tumor, but note that you should only label DCIS margins for info corresponding to the DCIS sample.

---

**LVI**

LVI should be marked as present if lymphovascular invasion is known to be present, absent if it is known to be absent, and Unknown otherwise. Note that we are combining both extensive and non-extensive lymphovascular invasion into a common category of "present". In the path note, you will generally find LVI in the 'Comments' section under the details of a specific specimen. Angiolymphatic invasion, lymphovascular invasion, lymphatic invasion, and vascular invasion are terms often used interchangeably among pathologists to describe the histologic finding of tumor cells within a vessel, and you should look out for either of these terms in the report.

---

**How to determine the 'worst' disease specimen?**

We want to ensure that the document level labels we select for a note correspond to the things we highlight in the note text. If a path report has both DCIS and invasive disease, the invasive disease is the 'worse' and all document level items and text highlights (unless explicitly related to DCIS) should refer to the invasive disease - i.e. grade, ER/PR/HER2 should refer to the invasive disease. Determining the 'worst disease' present can get tricky when the path note text has multiple path specimens with different types of invasive specimens. In general, if there are multiple invasive specimens, we aim to pick the 'worst' of these and have our text labels for this sample correspond to the labels at the document level. The 'worst' disease is tricky but general guides are that this will be the sample with the worst grade and/or size. If the path report has multiple samples but from different points in time (say from different institutions or before/after treatment), choose the sample that best reflects the patient's current disease state at the time of the note - so after treatment or the 'revised' impressions.

If there are two foci of invasive disease and they have different sizes – you should only highlight one of these two foci. Which of the two foci should be annotated depends on the analysis of the 'worst' disease – this is subjective but it'll generally be the larger of the two foci or the one with the worst grade.

### Section S2: GPT model prompts and settings

0613 version of the GPT-3.5-turbo model and 0314 version of the GPT-4 model were used via the Microsoft Azure OpenAI studio platform for all the experiments. The API version was 2023-05-15. The most deterministic temperature setting of 0 was used. The outputs were retrieved via the ChatCompletion API using the prompts described next. Additional prompt engineering or hyperparameter tuning was not performed.

**Prompts for extracting related pairs of entities for all the sub-tasks**

**System role:**

*Pretend you are a helpful Pathologist reading the given breast cancer pathology report."*

*Provide answers based on the pathology sample with the most aggressive or advanced cancer in the input report.*

*Do not use patient history to answer, only provide the current patient information as an answer.*

*Answer as concisely as possible in the given format.*

**User prompt template:**

*Provide the type of pathology report, biopsy procedure type, sites examined, sites of cancer, histological subtype, total number of lymph nodes involved, estrogen receptor status, progesterone receptor status, her2 gene amplification status, tumor grade, lympho-vascular invasion, final resection margins for invasive tumor, and final*

*resection margin for DCIS tumor. Notes about breast reconstruction surgery, or those unrelated to breast cancer are irrelevant here. Unknown option refers to the case where the answer cannot be inferred from the input note. For all irrelevant notes, return all everything other than path_type as the numeric option for Unknown. For molecular pathology report, report the path_type as the option for Unknown. Report the grade for treated tumors as Unknown, and do not report nuclear grade unless the most advanced tumor is of type DCIS. Numeric option for 'No malignancy' should be reported only if none of the samples is malignant. Numeric option for DCIS should always be reported as a histological subtype if it is present. Numeric option for 'Others' should be reported for histological subtype if a specific histological type is not discussed, but the tumor is not benign. For margins inference, if multiple margins have been reported, the margins after the final resection and associated with the worst prognosis, that is the one closest to the tumor, should be provided. Report DCIS margins as 'Unknown' if no DCIS tumor exists. Answer only with the most aggressive or advanced scenario in the current state. Do not use any history that has not been confirmed currently to answer.*

*Answer from the given options for each output:*

*pathology type: 1: Cytology 2. Histopathology 3. Either a report for breast reconstruction surgery, or a report unrelated to any breast cancer 4. Unknown.*

*biopsy procedure type: 1. Biopsy 2. Lumpectomy 3. Mastectomy 4. Unknown.*

*sites examined: 1. Left breast 2. Left lymph node 3. Other tissues than breast or lymph nodes 4. Right breast 5. Right lymph node 6. Unknown.*

*sites of cancer: 1. Left breast 2. Left lymph node 3. None 4. Other tissues than breast or lymph nodes 5. Right breast 6. Right lymph node 7. Unknown.*

*histological subtype: 1. DCIS 2. Invasive ductal carcinoma 3. Invasive lobular carcinoma 4. No malignancy was found 5. Other types of carcinoma than those mentioned 6. Unknown.*

*total number of lymph nodes involved: 1. 1 to 3 lymph nodes involved, 2. More than 10 lymph nodes involved 3. 4 to 9 lymph nodes involved 4. No lymph nodes involved 5. Unknown.*

*estrogen receptor status: 1. Negative 2. Positive 3. Unknown.*

*progesterone receptor status: 1. Negative 2. Positive 3. Unknown.*

*her2 gene amplification status: 1. Equivocal or indeterminate findings 2. Negative by FISH test 3. Positive by FISH test 4. Negative, but not with FISH test 5. Positive, but not with FISH test 6. Unknown.*

tumor grade: 1. 1 or low 2. 2 or intermediate 3: 3 or high, 4: Unknown.

lympho-vascular invasion: 1. Absent 2. Present 3. Unknown

final margins for invasive tumor: 1. Less than 2mm 2. More than or equal to 2mm 3. Positive margin 4. Unknown.

final resection margins for DCIS tumor: 1. Less than 2mm 2. More than or equal to 2mm 3. Positive margin 4. Unknown.

Provide the answers as a json in the following format, only using task-specific numeric options as specified above:

{

'path_type': option number for pathology type,

'biopsy': option number for biopsy procedure type,

'sites_examined': [list of all site option numbers that were examined for tumor],

'sites_cancer': [list of all site option numbers where cancer is found],

'histology': [list of all histological subtype option numbers for the most invasive tumor and DCIS],

'lymph_nodes_involved': option number for the group that includes the total number of lymph nodes involved,

'er': option number for estrogen receptor status,

'pr': option number for progesterone receptor status,

'her2': option number for her2 gene amplification status,

'grade': option number for tumor grade,

'lvi': option number for lympho-vascular invasion,

'margins': option number for final margins for invasive tumor,

'dcis_margins': option number for final margins for dcis tumor,

}

Do not provide answer as a list for anything except sites_examined, sites_cancer and histology.

### Section S3: Supervised classifier settings

For supervised classification models, the following settings were used:
English language stop words were removed from pathology reports before further processing. To find the best parameters for the random forests model, a random grid search was performed, using 3-fold cross-validation on the training data and 15 iterations. The explored parameters included the number of estimators, max depth, minimum samples split, minimum samples leaf, and maximum leaf nodes, as shown in **Table ST1**. n-grams in the range of [1–4] were tested to examine the impact of the length of the phrases on the model performance. The final model parameters were selected based on the best macro-average F1 score on the validation data. The final selected length of the ngrams, depending on the classification task, is provided in Table ST2.

**Table ST1.** Random grid search parameters for the random forests classifier.

|  | Range | | |
|---|---|---|---|
| **RF parameters** | minimum | maximum | **Number of steps** |
| number of estimators | 300 | 300 | N/A |
| max depth | None | 50 | 6 |
| minimum samples split | 2 | 100 | 5 |
| minimum leaf split | 1 | 10 | 4 |
| maximum leaf nodes | None | 40 | 20 |

**Table ST2. Random Forest final ngram count selected for each task**

| Task Name | n-grams | stopword removal (y/n) |
|---|---|---|
| Pathology Type | 2 | n |
| ER | 3 | n |
| HER2 | 3 | y |
| PR | 3 | n |
| LVI | 2 | y |
| Margins | 3 | n |
| DCIS margins | 3 | y |
| Biopsy | 3 | y |
| Grade | 3 | y |
| Lymph node involvement | 3 | n |
| Site Examined | 2 | n |
| Site Disease | 2 | y |
| Histology | 3 | y |

For the LSTM model, an architecture with attention was used. Max sequence length was set as 4600 to include all pathology reports in the dataset. Furthermore, hidden layer dimensions were finally selected to be 128, the model included 2 fully-connected layers, and was trained with a dropout value of 0.5 and the batch size of 16 for a total of 70 epochs. Adam optimizer with 1e-5 weight decay was used for model optimization. Learning rate of 5e-4 was set for the tasks of pathology type classification, grade classification, HER2 classification, and ER classification. For the tasks of classifying biopsy type, lymphovascular invasion, progesterone receptor status, margins and DCIS margins, the learning of 5e-3 was used. For all other tasks, the learning rate was set as 1e-3.

For the UCSF-BERT model, the of batch size 16 along with 2 grad accumulation steps was used, the learning rate was set to 2e-5, weight decay was set to 0., and the Adam optimizer was used with an epsilon of 1e-8, maximum gradient norm of 1.0, and the model was trained for 40 epochs.